\documentclass[11pt,a4paper]{article}
\pdfoutput=1

\usepackage{naacl2021}
\usepackage{times}
\usepackage{latexsym}
\usepackage{import}
\usepackage{xcolor}

\usepackage{array}
\usepackage{float}
\usepackage{graphicx}
\usepackage{makecell}
\usepackage{url}
\usepackage{hhline}
\usepackage{hyperref}
\usepackage{amssymb}
\usepackage{amsmath}

\usepackage{arydshln}
\newcommand*{\action}{\fontfamily{lmss}\selectfont}
\newcommand*{\token}{\fontfamily{cmtt}\selectfont}
\newcommand{\berttoken}[1]{{\tt\MakeUppercase{#1}}}

\newcommand{\livechat}{Expert Live Chat}
\newcommand{\tcom}{Cascading Dialogue Success}

\title{Action-Based Conversations Dataset: \\
A Corpus for Building More In-Depth Task-Oriented Dialogue Systems}

\author{Derek Chen$^{\dag}$, Howard Chen$^{\dag}$, 
    Yi Yang$^{\dag}$, Alex Lin$^{\dag}$, Zhou Yu$^{\ddag}$ \\
	$^\dag$ASAPP, New York, NY 10007 \\
	$^\ddag$Columbia University, NY \\
	\texttt{\small $^\dag$\{dchen,hchen,yyang,alin\}@asapp.com, zy2461@columbia.edu}
}

\date{}

\begin{document}
\maketitle
\begin{abstract}
Existing goal-oriented dialogue datasets focus mainly on identifying slots and values. However, customer support interactions in reality often involve agents following multi-step procedures derived from explicitly-defined company policies as well.  To study customer service dialogue systems in more realistic settings, we introduce the Action-Based Conversations Dataset (ABCD), a fully-labeled dataset with over 10K human-to-human dialogues containing 55 distinct user intents requiring unique \emph{sequences of actions} constrained by policies to achieve task success.

We propose two additional dialog tasks, Action State Tracking and \tcom, and establish a series of baselines involving large-scale, pre-trained language models on this dataset. Empirical results demonstrate that while more sophisticated networks outperform simpler models, a considerable gap (50.8\% absolute accuracy) still exists to reach human-level performance on ABCD. \footnote[1]{All code and data will be available at \href{https://github.com/asappresearch/abcd}{this location}.}

\end{abstract}

\section{Introduction}
\begin{figure}
  \includegraphics[width=\linewidth]{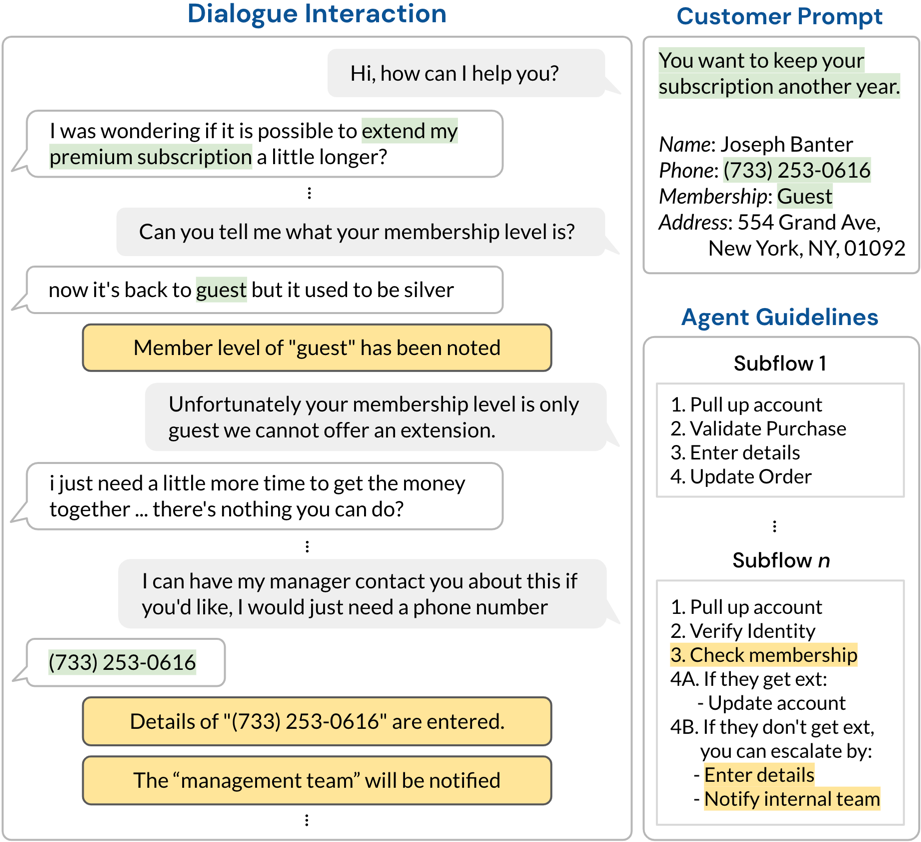}
  \caption{An interaction from ABCD (left) starts with the customer receiving a prompt (top right) to ground the dialogue. The agent follows the guidelines (bottom right) to identify the customer intent and to assist them in resolving the issue through a series of actions.} 
  \label{fig:abcd_fig1}
\end{figure}

The broad adoption of virtual assistants and customer service chatbots in recent years has been driven in no small part by the usefulness of these tools,
whereby \textit{actions} are taken on behalf of the user to accomplish their desired targets~\cite{alexa2019, google2019}.
Research into task-oriented dialogue has concurrently made tremendous progress on natural language understanding of user needs~\cite{wu2019transferable, rastogi2019towards, liang2019moss}.  However, selecting actions in real life requires not only obeying user requests, but also following practical policy limitations which may be at odds with those requests.  For example, while a user may ask for a refund on their purchase, an agent should only honor such a request if it is valid with regards to the store's return policy.  Described in actions, before an agent can {\action [Offer Refund]}, they must first {\action [Validate Purchase]}. 
Furthermore, resolving customer issues often concerns multiple actions completed in succession with a specific order since prior steps may influence future decision states. (See Figure~\ref{fig:abcd_fig1}) 

To more closely model real customer service agents, we present the Action-Based Conversations Dataset (ABCD) consisting of 10,042 conversations containing numerous actions with precise procedural requirements.  These actions differ from typical dialogue acts because tracking them necessitates striking a balance between external user requests and internally-imposed guidelines.  Thus, the major difference between ABCD and other dialogue datasets, such as MultiWOZ~\cite{budzianowski2018multiwoz}, is that it asks the agent to adhere to a set of policies while simultaneously dealing with customer requests.

While the prevalent data collection paradigm involves Wizard-of-Oz techniques, our situation containing asymmetric speakers compelled the design of a novel \livechat~system.  Our dataset includes asymmetric speakers because, unlike customers, agents must undergo extensive training to be able to navigate the Agent Guidelines during real-time conversations.  This makes a naive pairing process untenable since arbitrary matching might lead to chats containing two users who share the same role. 

Based on the unique aspects of ABCD, we propose two new tasks.  To start, Action State Tracking (AST) closely mirrors the format of Dialogue State Tracking where the user intent is inferred from the dialogue history.  AST then differs since the correct state must also be reconciled with the requirements outlined in the Agent Guidelines. As a second task, \tcom~(CDS) extends this notion across the entire conversation.  At each turn, the agent decides to take an action, respond with an utterance or end the chat.  As needed, the agent should also predict the right action or select the best utterance.

For each task, we build various models to establish baseline performance and to highlight the importance of each constraint.  Experiments show that in addition to conversation history, conditioning on the Agent Guidelines further boosts performance, with top models relying on both aspects to reach 31.9\% accuracy.  Additional results show removing action context hurts performance, implying the importance of taking into account the sequential nature of actions.  Lastly, human evaluation reaches 82.7\%, demonstrating ample room for future improvement.



The contribution of this work is three-fold:
(1)  We provide a novel, large-scale dataset containing context-dependent, procedural actions along with corresponding Agent Guidelines. (2) We establish a new technique called \livechat~for capturing natural dialogue between two unequal interlocutors. (3) We propose two metrics, Action State Tracking and \tcom, for measuring dialogue comprehension with policy constraints. Finally, we build on pretrained neural models to serve as baselines for these tasks.
\section{Related Work}

\paragraph{Traditional Dialogue Datasets}  In recent years, dialogue datasets have grown in size from hundreds of conversations to the tens of thousands~\cite{henderson2014second, budzianowski2018multiwoz, peskov2019multi}. 
Unlike open-domain chatbots often built for entertainment, task-oriented dialogue systems trained on such datasets are intended for solving user issues.   The resolution of these issues implicitly requires taking actions, where an action is a non-utterance decision that depends on both user and system inputs.  Despite the tremendous number of dialogues, examples in previous benchmarks fixate on the single knowledge base (KB) lookup action where the agent searches for an item that matches the user's desires and is available in the KB.
By sticking to this sole interaction, conversations can be generated through rules ~\cite{weston2015babi}, paraphrased from templates~\cite{byrne2019taskmaster} or taken from static text scenarios~\cite{zhang2018personalizing}, leading to dialogues that are predominantly homogeneous in nature.  

Many datasets have scaled to more domains as well~\cite{eric2017key, budzianowski2018multiwoz, peskov2019multi}
Since each new domain introduces a KB lookup requiring different slot-values, the number of unique actions grows as a linear function of the number of domains covered.  Rather than expanding wider, ABCD instead focuses deeper by increasing the count and diversity of actions within a single domain. 


\begin{table*}[!htp]
\small
\centering
\begin{tabular}{|m{1.1cm}|m{0.9\linewidth}|}
\hline
\textbf{Subflows} & recover-username,\textsuperscript{1} recover-password,\textsuperscript{1} reset-2fa,\textsuperscript{1}
status-service-added,\textsuperscript{2} status-service-removed,\textsuperscript{2} status-shipping-question,\textsuperscript{2} status-credit-missing,\textsuperscript{2} manage-change-address,\textsuperscript{2} manage-change-name,\textsuperscript{2} manage-change-phone,\textsuperscript{2} manage-payment-method,\textsuperscript{2}
status-mystery-fee,\textsuperscript{3} status-delivery-time,\textsuperscript{3} status-payment-method,\textsuperscript{3} status-quantity,\textsuperscript{3} manage-upgrade,\textsuperscript{3} manage-downgrade,\textsuperscript{3} manage-create,\textsuperscript{3} manage-cancel,\textsuperscript{3}
refund-initiate,\textsuperscript{4} refund-update,\textsuperscript{4} refund-status,\textsuperscript{4} return-stain,\textsuperscript{4} return-color,\textsuperscript{4} return-size,\textsuperscript{4} 
bad-price-competitor,\textsuperscript{5} bad-price-yesterday,\textsuperscript{5} out-of-stock-general,\textsuperscript{5} out-of-stock-one-item,\textsuperscript{5} promo-code-invalid,\textsuperscript{5} promo-code-out-of-date,\textsuperscript{5} mistimed-billing-already-returned,\textsuperscript{5} mistimed-billing-never-bought,\textsuperscript{5} 
status,\textsuperscript{6} manage,\textsuperscript{6} missing,\textsuperscript{6} cost,\textsuperscript{6} 
boots,\textsuperscript{7} shirt,\textsuperscript{7} jeans,\textsuperscript{7} jacket,\textsuperscript{7} 
pricing,\textsuperscript{8} membership,\textsuperscript{8} timing,\textsuperscript{8} policy,\textsuperscript{8} 
status-active,\textsuperscript{9} status-due-amount,\textsuperscript{9} status-due-date,\textsuperscript{9} manage-pay-bill,\textsuperscript{9} manage-extension,\textsuperscript{9} manage-dispute-bill,\textsuperscript{9}  
credit-card,\textsuperscript{10} shopping-cart,\textsuperscript{10} search-results,\textsuperscript{10} slow-speed\textsuperscript{10}
  \\
\hline
\textbf{Actions} & verify-identity, ask-the-oracle, validate-purchase, make-password, promo-code, subscription-status, offer-refund, make-purchase, record-reason, enter-details, shipping-status, update-order, pull-up-account, update-account, send-link, notify-team, membership, search-faq, try-again, log-out-in, instructions, search-jeans, search-shirt, search-boots, search-jacket, search-pricing, search-membership, search-timing, search-policy, select-faq\\
\hline
\end{tabular}
\caption{Full ontology of Agent Guidelines decomposable into high-level flows describing the overall category and subflows defining a unique set of intents. All actions are also shown. Upper script numeral indicates the flow that the subflow belongs to. 1: account access, 2: manage account, 3: order issue, 4: product defect, 5: purchase dispute, 6: shipping issue, 7: single item query, 8: storewide query, 9: subscription inquiry, 10: troubleshoot site} \label{tab:ontology}
\end{table*}

\paragraph{Exploring Other Avenues} Multiple aspects are explored by conversational datasets attempting to mimic reality. \newcite{rashkin2018towards} studies the ability of a dialogue model to handle empathy, while ~\newcite{zhou2018commonsense} focuses on commonsense reasoning.  Another approach is to augment dialogues with multi-modality including audio~\cite{castro2019towards} or visual~\cite{das2017visual} components.  Other researchers have explored grounding conversations with external data sources such as personas~\cite{zhang2018personalizing}, online reviews~\cite{ghazvininejad2018knowledge} or large knowledge bases~\cite{dinan2018wizard}.  Intricate dialogues can also appear when studying collaboration~\cite{he2017learning, kim2019codraw} or negotiation~\cite{lewis2017deal, he2018decoupling} which strongly encourage interaction with the other participant. In comparison, ABCD aims to make dialogue more realistic by considering distinct constraints from policies. 

\paragraph{Dialogues with Policies} Procedural actions following strict guidelines naturally emerge in dialogue research geared towards real-world applications. Hybrid Code Networks encode business logic through masking templates since various behaviors become nonsensical in certain situations~\cite{williams2017hybrid}.  Research from \newcite{moiseeva2020multipurpose} studies multi-purpose virtual assistants that attempt to distinguish among thirteen explicit actions.  The closest prior work to ABCD is the Schema Guided Dialogue (SGD) dataset, which contains dozens of API calls that can be interpreted as individual actions sending commands to a SQL engine~\cite{rastogi2019towards}.  The functionality of these actions is occasionally restricted to reflect constraints of real-life services.  The action restrictions within ABCD are made explicit by the Agent Guidelines manual.
\section{Action-Based Conversation Dataset}
In this section, we describe the task setting of ABCD by following along with the example dialog shown in Figure ~\ref{fig:abcd_fig1}.

\subsection{Customer}
During data collection, customers are given a simple prompt (such as ``You want to keep your subscription another year.") instead of step-by-step instructions, which reflects how real-world customers innately understand their own issue, but only have a rough idea of how to resolve said issue.
Accordingly, customers within ABCD remain oblivious towards what values apply to which actions, nor are they aware that actions exist in first place.  
This ambiguity forces the agent and customer to collaboratively uncover the correct latent intent through back and forth communication, naturally leading to longer dialogues. 

\subsection{Customer Service Agent}
Following the standard dialog setup, the agent starts by parsing the dialogue history to capture the customer intent, 
which in Figure~\ref{fig:abcd_fig1} is a subscription extension.  ABCD then diverges as the next step involves interpreting the Agent Guidelines, a document representing the internal policies of a 
company in the online retail domain (See Table~\ref{tab:ontology}).  Using the guidelines, the trained agent should find the one unique subflow corresponding to the customer intent. Each subflow in turn is defined by exactly one unique sequence of actions. 

While identifying a subflow may seem straightforward, information asymmetry prevents the customers from directly revealing the name of their intent. For example, a customer might inquire about the status of their recent purchase, but an agent has over a dozen different subflows related to order statuses, so selecting the right one suddenly becomes highly non-trivial.

In our case, the agent eventually figures out the correct subflow and begins to execute actions, which consists of recording values given by the customer, namely the customer's full name or account ID in order to {\action [Pull up Account]}.  As the third action, the guidelines instruct the agent to ask for the customer's membership level.  After the customer supplies this information, the agent enters the  ``guest'' value into the agent dashboard by clicking the {\action[Membership]} button.  Buttons have variable slots that may or may not need to be filled, depending on the context (See Table~\ref{tab:ontology} for a full list). Dialogue success demands that agents execute a chain of such actions in the right order with the right values,  while simultaneously engaging the customer in natural language conversation. 

There are three reasons that make carrying out a series of actions more difficult than the task lets on.  To start, the permitted actions in a given state are determined not only by Agent Guidelines, but also by the user's desire, which may be in conflict.  For example, the customer in Figure~\ref{fig:abcd_fig1} wanted to extend their subscription, but the guidelines prevented the agent from doing so.  Secondly, actions must be completed in order.  This \textit{procedural} requirement comes from the realization that completing actions out of order (or with missing steps) do not make sense in many real-world scenarios.  For example, it is critical to {\action [Verify Identity]} before resetting someone's password, not after. Finally, actions themselves induce stochastic outcomes, preventing agents from memorizing patterns of subflow resolution.  As an example, {\action [Ask the Oracle]} often determines if a customer complaint was valid.  In the case of a company error, the agent is compelled to immediately resolve the issue, whereas a misunderstanding made by the customer warrants a different set of responses.

\section{Data Collection Methodology}
This section outlines how we collect and annotate our dataset with context-dependent actions. 

\subsection{Agent Training}
Managing complex guidelines requires
filtering for top agents, which we do by certifying Mechanical Turk (MTurk) workers through an extensive 20-question quiz touching on all aspects of task completion. Keeping the bar high, we set a minimum threshold of 80\% accuracy of the quiz which resulted in a low 20\% pass rate. After passing the exam, we offered the answer key to agents which further improved understanding.  We also created short, 10-minute tutorial videos to showcase how to handle the most difficult aspects of the task.  A group chat app was also deployed to offer live feedback for agents, simulating how supervisors coach customer service representatives in real life.  Finally, we carefully designed an incentive structure that rewards agents for correctly identifying the user intent to encourage clarification behavior. (Appendix~\ref{sec:training} covers more details.)

\subsection{\livechat}
\fontdimen2\font=3pt
Rather than utilizing Wizard-of-Oz techniques (such as in MultiWOZ), we developed \livechat~which contains three unique aspects:
\fontdimen2\font=2.9pt
(1) Conversations are conducted continuously in real-time. (2) Users involved are not interchange-able. (3) Players are informed that all  participants are human -- no wizard behind the scenes.

\subsubsection{Synchronous Two-person Dialogue}
Normal human conversations occur in real-time, but coordinating multiple users in this manner is resource-intensive, so other datasets often employed workarounds to avoid this difficulty.  For example, other works have applied rules~\cite{bordes2016learning}, templates~\cite{byrne2019taskmaster} or paraphrasing ~\cite{shah2018building} to produce conversations.  Wizard-of-Oz (WoZ) techniques incorporate humans into the mix by allowing one of them to play the system role as a wizard behind the scenes~\cite{kelley1984iterative}. In particular, ~\cite{budzianowski2018multiwoz} decomposed dialogues into individual turns, where for each turn a new author is responsible for reading the context and generating the next plausible response.  Despite the time-consuming nature, some datasets have produced synchronous dialogues between two humans~\cite{lewis2017deal}.  However, the skill sets of ABCD workers are notably unequal, exacerbating the matching problem. 

\subsubsection{Pairing Users of Unequal Capability}
\livechat~matches a highly trained agent with a knowledgeable, yet otherwise average customer in real-time. Since the backgrounds are uneven, unlike other datasets with concurrent users~\cite{lewis2017deal, zhang2018personalizing, das2017learning}, incoming Turkers cannot simply be randomly assigned a role.  In other words, having twenty participants does not necessarily equate to ten conversations since it's possible that only a quarter of them are qualified as agents.  When such an imbalance inevitably arises, one group must wait until someone from the other side becomes available. However, leaving either side waiting for too long leads to serious consequences since idle time directly affects their pay rate.

To minimize the likelihood of such an outcome, we first ensure that a reasonable pool of agents are always available.  Then, we increase the number of active customers by methodically inviting a subset of customers one batch at a time.  To do so, we established a qualification exam for customers to ensure their availability during a specified time period. Finally, we also redesigned the chat application to make the waiting room experience more palatable. (See Appendix~\ref{sec:pairing} for full breakdown.) With these changes, we successfully increased the pairing rate from 18 out of 80 active users up to 72 out of 83, an increase of nearly 400\%, while maintaining wait times under 10 minutes.

\begin{figure}
  \includegraphics[width=\linewidth]{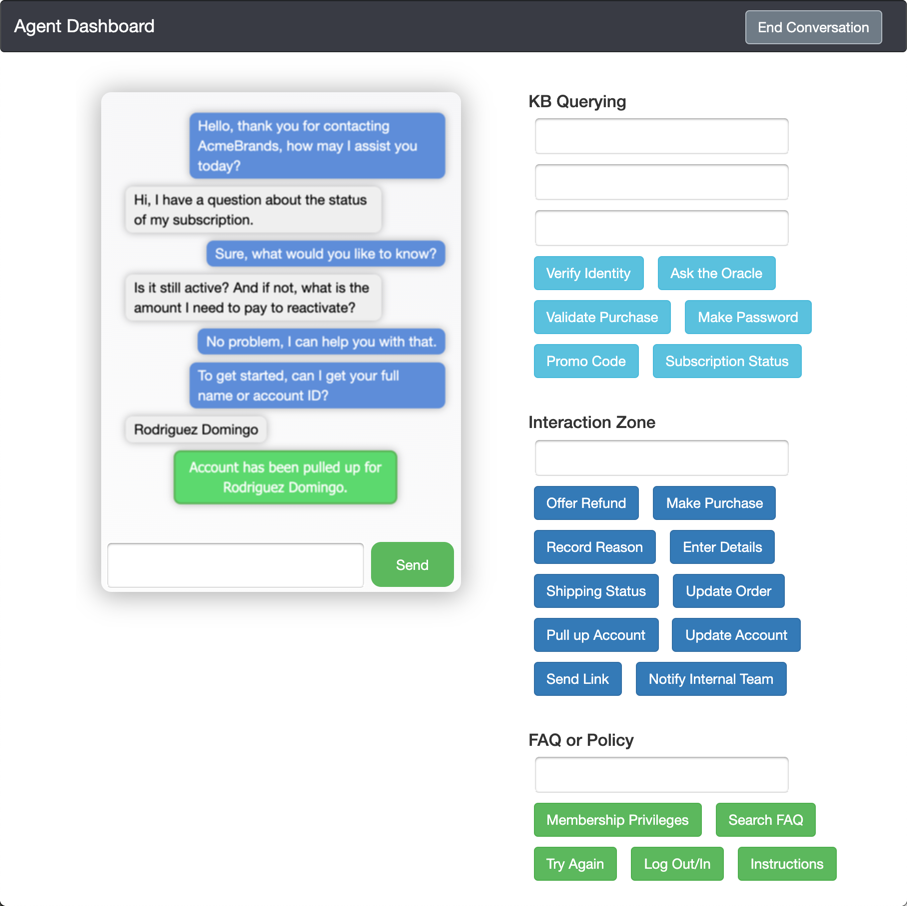}
  \caption{The Agent Dashboard is split into three sections.  KB Query actions always have system output, while actions in the Interaction Zone require user input. The FAQ/Policy section is associated with describing company policies and technical troubleshooting.}
  \label{fig:dashboard}
\end{figure}

\subsubsection{Interaction Framework}
Besides pairing, we increased the likelihood of collecting rich dialogues without the need for extensive instructions by optimizing the chat experience itself.  In particular, we observed the greatest gains by grounding the conversation to the relatable scenario of online shopping, which provided immediate context to participants without requiring any extra training.

For example, the \textit{Agent Dashboard} was arranged to closely reflect actual agent workspaces (Figure~\ref{fig:dashboard}). On the customer side, scenarios in the \textit{Customer Panel} included an image of the product being discussed, along with other meta-data such as the brand or price to match a true shopping experience as much as possible (Appendix~\ref{sec:customer}).  We also explicitly told customers the other speaker was human to encourage natural responses over confined commands meant for machines.  Most importantly, customers were given dynamically generated, natural-language prompts that did \textit{not} include information about the values needed to resolve their issue. As a general framework, \livechat~can be applied in any real-world scenario involving an expert and novice.  Indeed, increasing the verisimilitude of the experience is precisely what allowed higher quality dialogues to be generated by the workers.


\subsection{Annotation of Actions and Values}
The flows and subflows are automatically annotated since we have the provenance of each intent when generating the customer prompt.  Additionally, given the ground truth subflow of each conversation, we can deterministically map them to the correct section within the Agent Guidelines outlining the correct actions.  Calculating accuracy then becomes a simple exercise to align the predicted actions with the ones required by the manual.  In this way, we capture a key benefit of machine-generated text~\cite{shah2018building} without sacrificing the benefit of engaging real users. 

\begin{table*}
\centering
\resizebox{\textwidth}{!}{%
\begin{tabular}{lccccccccc}
\Xhline{1pt}
\textbf{Metric} & \textbf{DSTC2} & \textbf{M2M} & \textbf{KVRET} & \textbf{MultiWOZ} &  \textbf{SGD} & \textbf{MultiDoGO} & \textbf{ABCD}  \\
\hline
Num of Dialogues & 1,612 & 1,500 & 2,425 & 8,438 & 16,142 & \textbf{40,576} & 8,034 \\
Num of Turns & 23,354 & 14,796 & 12,732 & 113,556  & 329,964 & \textbf{813,834} & 177,407 \\
Num of Tokens & 199,431 & 121,977 & 102,077 & 1,490,615 & 3,217,369 & \textbf{9,901,235} & 1,626,160 \\
Avg. Turns / Dialogue & 14.49 & 9.86 & 5.25 & 13.46 & 20.44 & 20.06 & \textbf{22.08} \\
Avg. Tokens / Turn & 8.54 & 8.24 & 8.02 & \textbf{13.13} & 9.75 & 12.16  & 9.17 \\
Std Dev. Tokens / Turn & 2.95 & 5.99 & 6.07 & 6.19 & 6.48 & --* & \textbf{6.80} \\
Avg. Actions / Dialogue & 1.0 & 1.0 & 1.0 & 1.81 & 1.24 & --* & \textbf{3.73} \\
No. Unique Tokens & 986 & 1,008 & 2,842 & 23,689 & 30,352 & \textbf{70,003}  & 23,686\\
No. Unique Slots & 8 & 14 & 13 & 24 & 214 & 73  & \textbf{231} \\
No. Slot Values & 212 & 138 & 1,363 & 4,510 & 14,139 & \textbf{55,816} & 12,047 \\
No. Domains & 1 & 2 & 3 & 7 & 16 & 6 & \textbf{30}\\
\Xhline{1pt}
\end{tabular}}
\caption{Comparison of ABCD to similar dialogue datasets. Numbers reported are for the train split on all datasets, with bold values indicating the top score for each metric. *MultiDoGO is not public, unable to calculate new stats.} \label{tab:data-stats}
\end{table*}

\section{Dataset Statistics and Analysis}
We validate all dialogues to pass quality thresholds such as including a minimum number of actions and avoiding copy/paste behavior.  After filtering, we end up with 10,042 total conversations with an average of 22.1 turns -- the highest turn count among all compared datasets.  Unsurprisingly, ABCD includes more actions per dialogue than other datasets, by at least a factor of two.  ABCD also 
contains a lower absolute number of tokens, but also has the highest variance in the number of tokens per turn. (See Table \ref{tab:data-stats}.)

Since each subflow represents a unique customer intent, ABCD contains 55 user intents evenly distributed through the dataset. By interpreting buttons as domains, the dataset contains 30 domains and 231 associated slots, compared to 7 domains and 24 slots within MultiWOZ~\cite{budzianowski2018multiwoz}.

By grounding to the relatable scenario of chatting with customer support of an online retail company, speakers often showcase various forms of natural dialogue, such as offering diverse reasons for shopping or asking detailed follow-up questions.  Furthermore, the unconstrained nature of \livechat~allows users to chat with each other in a free-form style.  Dialogues exhibited normal texting behavior such as users speaking for many turns in a row or fixing typos with a star in the subsequent line.  Other examples of linguistic phenomenon can be observed in Table~\ref{tab:lang-phenom}.

\section{ABCD as a Dialogue Benchmark}
The novel features in ABCD brings two new dialog tasks, Action State Tracking and \tcom. We also build baseline systems that are variants of standard dialogue models and report their results on ABCD.

\subsection{Action State Tracking} 
Action State Tracking (AST) aims at detecting the pertinent intent by interpreting customer utterances while taking into account constraints from the Agent Guidelines, an aspect not considered in traditional dialog state tracking (DST). For example, a conceivable dialogue task might entail helping a customer {\action[Reset Password]} once this intent has been identified.  In contrast, the appropriate next step within AST is governed by the Agent Guidelines, which might require {\action[Verify Identity]} of the customer first, or any number of other actions, before executing the password reset.

Each series of actions is considered a unique subflow that belongs to a number of high-level conversational flows.  Each individual action includes the active button $b$ to click and its corresponding slots $s$ and values $v$.  The task consists of executing an action, which constitutes a single agent turn.  
More specifically, given a context $C_t = [x_1, x_2, \ldots, x_t]$ where $x_t$ can be a customer utterance $x^c_t$, an agent utterance $x^a_t$, or a prior action $x^b_t$, a model should predict the button of the current action as well as the relevant slots and values, if any exist $\{x^b_{t+1} = (b, s, v) \in \mathcal{B}\times \mathcal{S}\times \mathcal{V}\}$.

This structure is designed to mimic DST where each user intent is broken down into domains, slots and values $(d,s,v)$.  For both AST and DST, the higher level domain or button can have varying slots.  The reverse is also true -- a given slot can be associated with multiple domains or buttons.  Lastly, both contain values that can be enumerable (i.e. payment types or shipping statuses) or non-enumerable (phone numbers or email addresses). Following the pattern set by~\newcite{rastogi2019towards}, enumerable values are given in the ontology to be accessible by a model, whereas the non-enumerable items are not.

Despite the similar structure, AST deviates from DST since predicting the right action requires not only parsing the customer utterance, but also adhering to Agent Guidelines.  Suppose a customer is entitled to a discount which will be offered by issuing a {\action[Promo Code]}.  The customer might request 30\% off, but the guidelines stipulate only 15\% is permitted, which would make ``30'' a reasonable, but ultimately flawed slot-value.  To measure a model's ability to comprehend such nuanced situations, we adopt overall accuracy as the evaluation metric for AST. 

\subsection{\tcom}
Since the appropriate action often depends on the situation, we propose the \tcom~(CDS) task to measure a model's ability to understand actions in context.  Whereas AST assumes an action occurs in the current turn, CDS gives an agent the additional options of responding with an utterance or ending the conversation.  Moreover, proficiency is no longer measured as success over \textit{isolated} turns but rather as success over sequences of \textit{consecutive} turns.

Formally, given $C_t = [x_1, x_2, \ldots, x_t]$ as a context composed of utterances $x^c, x^a \in \mathcal{U}$ and actions $x^b \in \mathcal{A}$, a model should predict all remaining steps $x_{>t}$ along with their realized forms.  Possible next steps are to take an action, respond with text or end the task. When the next step is an action $x^b_{t+1}$, the model should predict the button with its slots and values as in AST.  If the agent speaks in the next step $x^a_{t+1}$, the model should rank the true utterance highest, as measured by recall metrics.\footnote{Sentences with similar semantics may be formulated in several ways, so we opt for response retrieval over text generation since common metrics (i.e. BLEU score) tend to become unreliable in these situations~\cite{liu2016not}.}  Finally, the model should recognize when to end the conversation. 

Rewarding the model only when it predicts every step correctly is 
counter-productive because minor variations in sentence order do not alter overall customer satisfaction.  Therefore, CDS is scored using a variation on \textit{Cascading Evaluation}~\cite{suhr2019executing}.  Rather than receiving a single score for each conversation, cascaded evaluation allows the model to receive ``partial credit'' whenever it successfully predicts each successive step in the chat.  This score is calculated on every turn, and the model is evaluated based on the percent of remaining steps correctly predicted, averaged across all available turns. (See Appendix~\ref{sec:cascade} for more details.)

\subsection{Baseline Models}
We also run several baselines on these new tasks. The backbone of all our baseline systems is a pre-trained Transformer-based model acting as a context encoder. More specifically, given the dialogue history as a series of utterances, we first join the utterances together with a {\token [SEP]} token and then tokenize the entire input using WordPiece~\cite{schuster2012wordpiece}. Next, we feed the entire input into a BERT model and perform a learned pooling on the hidden states in the final layer, which results in a fixed-length latent vector $~h_{enc} \in \mathbb{R}^{128}$~\cite{wolf2019transfertransfo}. Afterwards, we attach a variety of prediction heads conditioned on the $h_{enc}$ vector to generate the final output.  Details of the prediction heads for the two proposed tasks are described next.

We break down Action State Tracking (AST) into two sub-problems, button-slot prediction and value-filling. Given the ontology, button prediction is a straightforward classification task over 231 known options, so the prediction head is just a linear classifier with a softmax activation for normalization: $P_{b\cdot slot} = \text{Softmax}(W_a h_{enc}^{\top} + b_a)$.

To handle value-filling, we further decompose the task into predicting enumerable and non-enumerable values.  The ontology lists out all $|E|$ enumerable values, so the prediction head $p^{enum}$ simply maps the hidden state $h_{enc}$ into the appropriate dimensions.  To handle non-enumerable values, we follow the insight from ~\cite{ma2019end} which notes that practically all such values are stated by the customer in conversation, so a model can copy these values from the tokenized context.  During pre-processing, we extract up to $|N|$ unique tokens from the natural language customer utterances, where $p^{copy}$ then represents the distribution over these possible options.\footnote{Choosing larger $|N|$ leads to higher recall, but lower precision.  We found $N=100$ to work well in practice.}

We imitate the TRADE architecture from  ~\cite{wu2019transferable}, where conditioned on the action, the model chooses to either copy from the context $p_{copy}$ or select from the enumerable entities $p_{enum}$ based on a gating mechanism.  The gate is conditioned on the hidden state $h_{enc}$ as well as a learned context vector $c_i$. Concretely,
\begin{align*} 
    p^{enum} &= \text{Softmax}(W_e h_{enc}^{\top} + b_e) \in \mathbb{R}^{|E|} \\
    p^{copy} &= \text{Softmax}(W_c h_{enc}^{\top} + b_c) \in \mathbb{R}^{|N|}  \\
	c_i      &= W_c^{\top} \cdot p^{copy} \in \mathbb{R}^{hid} \\
	p^{gate} &= \sigma(W_g \cdot [h_{enc}; c_i ]) \in \mathbb{R}^1 \\
	P_{val} = [p^{gate} &\times p^{copy} ; (1 - p^{gate}) \times p^{enum}] \in \mathbb{R}^{|E+N|}
\end {align*}
where $\sigma$ represents the Sigmoid function and $[\cdot;\cdot]$ is the concatenation operation.  The final value predictions are the argmax of $P_{val}$ which merge the probabilities of $p^{enum}$ and $p^{copy}$ together.

For \tcom~(CDS), we also tackle next step selection, utterance ranking, and intent classification.  Next step selection is a choice between \textit{retrieve utterance}, \textit{take action} and \textit{end conversation}.  Intent classification consists of choosing from the 55 available subflows.  Given this basic setting, both tasks use the same setup of a linear layer followed by a softmax, albeit with their own respective weights $W_{NS} \in \mathbb{R}^{3 \times hid}$ and $W_{IC} \in \mathbb{R}^{55 \times hid}$.  When the next step is to \textit{take action}, the AST model is reused to determine the button-slot and value.  When \textit{end conversation} is selected, all future predictions are ignored, much like an 
{\token <EOS>}
symbol signifies stopping.

This leaves us with utterance ranking, which is only evaluated when \textit{retrieve utterance} is chosen as the next step. Our ranker reproduces the design from~\cite{guu2020realm}, where the encoded context  $h_{ctx}$ is compared against each encoded candidate response $h_{cand}$ to produce a ranking score. To embed each $j^{th}$ candidate $d_j$ we first create its input $d_j^{input}$. Following standard practice, we prepend the candidate text $d_j$ with {\token [CLS]}, separate the individual utterances  $u_i$ within the candidate response using a {\token [SEP]} token, and append a final {\token [SEP]} token afterwards. ~\cite{devlin2018bert}. This input $d_j^{input}$ is then fed into a static pretrained BERT model to get an initial hidden state, which is finally projected using a learned weight $W_{d_j} \in \mathbb{R}^{128 \times hid}$ to produce $h_{cand}$.  To obtain $h_{ctx}$ we start with the hidden state $h_{enc}$ from before and apply a projection matrix $W_{UR} \in \mathbb{R}^{128 \times hid}$ to reach the desired dimensionality. 
\begin{align*}
	d_j^{\rm input} &=  \berttoken{[CLS]} u_1 \berttoken{[SEP]} u_2 \berttoken{[SEP]} ... \berttoken{[SEP]} u_n \berttoken{[SEP]} \\
	h_{cand} &= W_{d_j}  \text{BERT}_{\text{base}}(d_j^{input})^{\top} \in \mathbb{R}^{128} \\
	h_{ctx} &= W_{UR}\: h_{enc}^{\top} \in \mathbb{R}^{128} \\
	f(x_i, d_j) &= h_{ctx}^{\top}\: h_{\rm cand} \\
	P^{\rm rank}_j &= \frac{\text{exp}(f(x_i, d_j))}{\Sigma_{d_j'} \exp f(x_i, d_j')}
\end{align*}
The final rank is given by normalizing each $j^{th}$ score against all other candidate scores.
We use the training objective from ~\cite{henderson2019training} to calculate the loss:
$$  \mathbb{J} = \sum_{j=1}^{M=100}P(x_i, d_j) - \sum_{i=1}^{M} \log \sum_{j=1}^{M}\exp^{f(x_i, d_j)} $$
where $M$ is the size of the total candidate set.

\subsection{Experiments}
We performed experiments on the two newly proposed tasks, AST and CDS.  AST consists of two subtasks, button-slot prediction and value-filling, while CDS builds on this with three additional subtasks of next step selection, utterance ranking, and intent classification. For both tasks, we experimented with two types of frameworks, a pipeline version and an end-to-end version. The pipeline version trains each subtask separately while the end-to-end optimizes all tasks jointly~\cite{liang2019moss, rastogi2020schema, ham2020end}.

The pipeline model uses a BERT model trained with the RAdam optimizer~\cite{liu2019radam}.
To test the performance of different pretrained models under the end-to-end framework, we experiment with three additional encoders, AlBERT~\cite{lan2019albert}, RoBERTa~\cite{liu2019roberta} and RoBERTa-Large.  AlBERT model has an inter-sentence coherence task and a lighter memory footprint compared to BERT, while RoBERTa model has substantially more data and hyper-parameter tuning in pretraining than BERT. In the future, we also plan to include GPT-based models, such as DialoGPT~\cite{zhang2019dialogpt} in our comparison. 

\subsection{Results}
For both tasks, moving from the pipeline architecture to a jointly trained method displayed noticeable improvement in accuracy.  As hinted at in prior works~\cite{liang2019moss}, we suspect the group effort gives each subtask extra supervision from other subtasks for more data efficient training. 
In the AST task, we found steady improvements as we move from the older to the newer models with vanilla BERT at 59.5\% accuracy and RoBERTa doing the best at 65.8\%.  For the CDS task, we found a similar trend where RoBERTa-Large outperforms BERT, but only by a mere 0.6\%.  We hypothesize this small gap between models is due to the fact that none were particularly trained on dialogue data which impacts their ability to produce a useful encoding~\cite{wu2020probing}.

Separately, we evaluate CDS subtask difficulty by asking human volunteers to select the correct label from a list of possible options. As an example, workers would be presented with 55 different classes for Intent Classification and asked to choose the right one. Since humans typically struggle when choosing from large collections of items, fine-tuned models performed roughly on par or better compared to humans in this unnatural setting.  On the other hand, human evaluation for the overall CDS task was judged by measuring the success rate in a standard conversational scenarios where behavioral instincts are activated, so humans were able to excel on this environment.


\begin{table}
\centering
\resizebox{\linewidth}{!}{%
\begin{tabular}{l@{\hspace{0.5em}}cccc}
\Xhline{1pt}
\textbf{Metric} & \textit{\textbf{Pipeline}} & \textit{\textbf{BERT}} & \textit{\textbf{AlBERT}} & \textit{\textbf{RoBERTa}} \\
\hline
B-Slot          & 86.7\% & 89.9\% & 90.9\% & \textbf{93.6\%} \\
Value           & 42.1\% & 61.6\% & 61.0\% & \textbf{67.2\%} \\
Action          & 32.3\% & 59.5\% & 59.2\% & \textbf{65.8\%} \\
\Xhline{1pt}
\end{tabular}}
\caption{Metrics for Action-State Tracking. Pipeline values come from models trained on individual subtasks, other models are trained jointly end-to-end.} \label{tab:action-pipe}
\end{table}

\begin{table*}
\centering
\resizebox{0.9\textwidth}{!}{%
\begin{tabular}{|l|cccccc|}
\Xhline{1pt}
\textbf{Model} & \textbf{Intent} & \textbf{Nextstep} & \textbf{B-Slot} & \textbf{Value} & \textbf{Recall@1/5/10} & \textbf{Cascading Eval}\\
\hhline{|=|======|}
Human & 85.5\%   & 84.0\% & 79.0\% & 77.5\% &        N/A      & 82.7\% \\

Pipeline       & 90.4\%  & 83.8\%  & 86.7\% & 42.1\% & 26.2/51.7/63.1 & 18.2\%    \\
\hline
BERT-base      & 89.3\%  & 87.6\%  & 85.9\% & 73.1\% & 21.7/46.6/58.7 & 31.3\%    \\
AlBERT         & 88.5\%  & 87.2\%  & 86.1\% & 70.4\% & \textbf{22.1}/47.4/58.9 & 31.2\%    \\
RoBERTa        & 89.7\%  & \textbf{87.8\%}  & 87.6\% & 73.1\% & 21.6/46.7/58.6 & 31.5\%    \\
RoBERTa-Large  & \textbf{90.5\%}  & 87.5\%  & \textbf{88.5\%} & \textbf{73.3\%} & 22.0/\textbf{47.8}/\textbf{59.1} & \textbf{31.9\%}    \\
\hline
BERT-base w/o Action Info  & 88.4\%  & 76.8\% & 83.7\% & 63.4\% & 18.6/43.0/57.9 & 29.2\% \\
\hdashline
BERT-base w/ Guidelines     & 83.2\%  & 87.5\% & 85.6\% & 72.4\% & 21.8/46.9/58.5 & 30.6\% \\
BERT-base w/ Intent Info    & 100\%   & 88.6\% & 88.9\% & 73.8\% & 22.2/47.6/59.1 & 32.3\% \\
BERT-base w/ Intent + Guide & 100\%   & 89.2\% & 89.3\% & 74.0\% & 22.6/48.1/59.4 & 32.7\% \\
\Xhline{1pt}
\end{tabular}}
\caption{Cascading dialogue success task performance with breakdown of all five subtasks.  Numbers displayed are the average of three seeds.  Human evaluation conducted with size of 100 samples per person.} \label{tab:tcom}
\end{table*}

\subsection{Ablation Study}
We perform an ablation study to test the significance of the key features in ABCD. Recall, actions are characterized by their dual nature of requiring signals from both the customer and the company guidelines. To that end, we provided the ground truth intent to measure the impact of the customer side. Conversely, we also test the company side by masking out invalid buttons based on the insight that the Agent Guidelines are useful for narrowing down the range of possible actions. In both situations, we would expect that providing such oracle guidance would boost performance.  Lastly, note that the appropriate action depends on the outcomes of prior actions, so for a final experiment we removed prior actions and their explanations from the context to test their impact on task success. (See Appendix~\ref{sec:info} for details.)

We observe that supplying the intent information to the BERT model causes a noticeable boost in dialog success, bringing the score to 32.3\%.  However, augmenting the model with knowledge of the guidelines unexpectedly dropped performance down to 30.6\%.  Further analysis revealed the imperfect intent classifier would occasionally mask out valid buttons, leaving only incorrect ones to choose from.  As a result, the downstream action predictor would be prevented from doing its job, causing errors to accumulate.  To test this hypothesis, we ran another model (Intent+Guide) which had access to guidelines along with an oracle intent classifier.  This model reached the peak observed performance of 32.7\%, highlighting the importance of both components. As a final result, removing action information away from action-based conversations unsurprisingly causes a major
performance drop (Table~\ref{tab:tcom}).

\section{Conclusion and Future Work}
In conclusion, we have presented ABCD which includes over 10K dialogues that incorporate procedural, dual-constrained actions.  Additionally, we established a scalable method for collecting live human conversations with unequal partners.  We found that pre-trained models perform decent on Action State Tracking, but there is a large gap between humans agents and the top systems for \tcom.

We plan to incorporate GPT-related models~\cite{hosseini2020simple}, as alternate forms of preprocessing have shown promise in other NLP tasks.  Other techniques could also be used to incorporate speaker info, action semantics and other meta-data.
Wholly new systems that attend to the Agent Guidelines in a fully differentiable manner are also worth exploring. By grounding dialogues to in-depth scenarios with explicit policies, we hope to have pushed towards a better understanding of dialogue success. 




\section*{Acknowledgments}
The authors would like to thank Tao Lei, Felix Wu and Amnol Kabra for their feedback and support.  We would also like to thank the anonymous NAACL 2021 reviewers for pointing out specific areas of confusion in our submission, which we have tried our best to clarify.

\section*{Ethical Considerations}
This paper presents a new dataset which was collected through the use of crowdworkers.  All agent workers were compensated a fair wage based on their local standard of living, where their location was determined during the vetting process.  (Please refer to Appendix~\ref{sec:training} for more details.)

\bibliography{naacl2021}

\begin{thebibliography}{44}
\expandafter\ifx\csname natexlab\endcsname\relax\def\natexlab#1{#1}\fi

\bibitem[{Amazon(2019)}]{alexa2019}
Amazon. 2019.
\newblock \href {https://developer.amazon.com/en-US/alexa/alexa-skills-kit}
  {\emph{Alexa Skills Kit}}.

\bibitem[{Bordes et~al.(2017)Bordes, Boureau, and Weston}]{bordes2016learning}
Antoine Bordes, Y-Lan Boureau, and Jason Weston. 2017.
\newblock Learning end-to-end goal-oriented dialog.
\newblock In \emph{ICLR}. OpenReview.net.

\bibitem[{Budzianowski et~al.(2018)Budzianowski, Wen, Tseng, Casanueva, Ultes,
  Ramadan, and Gasic}]{budzianowski2018multiwoz}
Pawel Budzianowski, Tsung-Hsien Wen, Bo-Hsiang Tseng, I{\~n}igo Casanueva,
  Stefan Ultes, Osman Ramadan, and Milica Gasic. 2018.
\newblock Multiwoz - a large-scale multi-domain wizard-of-oz dataset for
  task-oriented dialogue modelling.
\newblock In \emph{Proceedings of the 2018 Conference on Empirical Methods in
  Natural Language Processing, Brussels, Belgium, October 31 - November 4,
  2018}, pages 5016--5026. Association for Computational Linguistics.

\bibitem[{Byrne et~al.(2019)Byrne, Krishnamoorthi, Sankar, Neelakantan,
  Goodrich, Duckworth, Yavuz, Dubey, Kim, and Cedilnik}]{byrne2019taskmaster}
Bill Byrne, Karthik Krishnamoorthi, Chinnadhurai Sankar, Arvind Neelakantan,
  Ben Goodrich, Daniel Duckworth, Semih Yavuz, Amit Dubey, Kyu-Young Kim, and
  Andy Cedilnik. 2019.
\newblock Taskmaster-1: Toward a realistic and diverse dialog dataset.
\newblock In \emph{Proceedings of the 2019 Conference on Empirical Methods in
  Natural Language Processing and the 9th International Joint Conference on
  Natural Language Processing, EMNLP-IJCNLP 2019, Hong Kong, China, November
  3-7, 2019}, pages 4515--4524. Association for Computational Linguistics.

\bibitem[{Castro et~al.(2019)Castro, Hazarika, P{\'e}rez-Rosas, Zimmermann,
  Mihalcea, and Poria}]{castro2019towards}
Santiago Castro, Devamanyu Hazarika, Ver{\'o}nica P{\'e}rez-Rosas, Roger
  Zimmermann, Rada Mihalcea, and Soujanya Poria. 2019.
\newblock Towards multimodal sarcasm detection (an \_obviously\_ perfect
  paper).
\newblock In \emph{Proceedings of the 57th Conference of the Association for
  Computational Linguistics, ACL 2019, Florence, Italy, July 28- August 2,
  2019, Volume 1: Long Papers}, pages 4619--4629. Association for Computational
  Linguistics.

\bibitem[{Das et~al.(2017{\natexlab{a}})Das, Kottur, Gupta, Singh, Yadav,
  Moura, Parikh, and Batra}]{das2017visual}
Abhishek Das, Satwik Kottur, Khushi Gupta, Avi Singh, Deshraj Yadav,
  Jos{\'e}~MF Moura, Devi Parikh, and Dhruv Batra. 2017{\natexlab{a}}.
\newblock Visual dialog.
\newblock In \emph{Proceedings of the IEEE Conference on Computer Vision and
  Pattern Recognition}, pages 326--335.

\bibitem[{Das et~al.(2017{\natexlab{b}})Das, Kottur, Moura, Lee, and
  Batra}]{das2017learning}
Abhishek Das, Satwik Kottur, Jos{\'e}~MF Moura, Stefan Lee, and Dhruv Batra.
  2017{\natexlab{b}}.
\newblock Learning cooperative visual dialog agents with deep reinforcement
  learning.
\newblock In \emph{Proceedings of the IEEE International Conference on Computer
  Vision}, pages 2951--2960.

\bibitem[{Devlin et~al.(2019)Devlin, Chang, Lee, and
  Toutanova}]{devlin2018bert}
Jacob Devlin, Ming-Wei Chang, Kenton Lee, and Kristina Toutanova. 2019.
\newblock Bert: Pre-training of deep bidirectional transformers for language
  understanding.
\newblock In \emph{Proceedings of the 2019 Conference of the North American
  Chapter of the Association for Computational Linguistics: Human Language
  Technologies, NAACL-HLT 2019, Minneapolis, MN, USA, June 2-7, 2019, Volume 1
  (Long and Short Papers)}, pages 4171--4186. Association for Computational
  Linguistics.

\bibitem[{Dinan et~al.(2019)Dinan, Roller, 0001, Fan, Auli, and
  Weston}]{dinan2018wizard}
Emily Dinan, Stephen Roller, Kurt~Shuster 0001, Angela Fan, Michael Auli, and
  Jason Weston. 2019.
\newblock Wizard of wikipedia: Knowledge-powered conversational agents.
\newblock In \emph{ICLR}. OpenReview.net.

\bibitem[{Eric et~al.(2017)Eric, Krishnan, Charette, and Manning}]{eric2017key}
Mihail Eric, Lakshmi Krishnan, Francois Charette, and Christopher~D. Manning.
  2017.
\newblock Key-value retrieval networks for task-oriented dialogue.
\newblock In \emph{SIGDIAL Conference}, pages 37--49. Association for
  Computational Linguistics.

\bibitem[{Ghazvininejad et~al.(2018)Ghazvininejad, Brockett, Chang, Dolan, Gao,
  Yih, and Galley}]{ghazvininejad2018knowledge}
Marjan Ghazvininejad, Chris Brockett, Ming-Wei Chang, Bill Dolan, Jianfeng Gao,
  Wen-tau Yih, and Michel Galley. 2018.
\newblock A knowledge-grounded neural conversation model.
\newblock In \emph{Thirty-Second AAAI Conference on Artificial Intelligence}.

\bibitem[{Google(2019)}]{google2019}
Google. 2019.
\newblock \href {https://developers.google.com/actions/overview} {\emph{Actions
  on Google Assistant}}.

\bibitem[{Guu et~al.(2020)Guu, Lee, Tung, Pasupat, and Chang}]{guu2020realm}
Kelvin Guu, Kenton Lee, Zora Tung, Panupong Pasupat, and Ming-Wei Chang. 2020.
\newblock Realm: Retrieval-augmented language model pre-training.
\newblock \emph{CoRR}, abs/2002.08909.

\bibitem[{Ham et~al.(2020)Ham, Lee, Jang, and Kim}]{ham2020end}
Donghoon Ham, Jeong-Gwan Lee, Youngsoo Jang, and Kee-Eung Kim. 2020.
\newblock End-to-end neural pipeline for goal-oriented dialogue system using
  gpt-2.
\newblock In \emph{Proc. of the 34th {AAAI} Conference on Artificial
  Intelligence}. ACL.

\bibitem[{He et~al.(2017)He, Balakrishnan, Eric, and Liang}]{he2017learning}
He~He, Anusha Balakrishnan, Mihail Eric, and Percy Liang. 2017.
\newblock Learning symmetric collaborative dialogue agents with dynamic
  knowledge graph embeddings.
\newblock In \emph{Proceedings of the 55th Annual Meeting of the Association
  for Computational Linguistics, ACL 2017, Vancouver, Canada, July 30 - August
  4, Volume 1: Long Papers}, pages 1766--1776. Association for Computational
  Linguistics.

\bibitem[{He et~al.(2018)He, Chen, Balakrishnan, and Liang}]{he2018decoupling}
He~He, Derek Chen, Anusha Balakrishnan, and Percy Liang. 2018.
\newblock Decoupling strategy and generation in negotiation dialogues.
\newblock In \emph{Proceedings of the 2018 Conference on Empirical Methods in
  Natural Language Processing, Brussels, Belgium, October 31 - November 4,
  2018}, pages 2333--2343. Association for Computational Linguistics.

\bibitem[{Henderson et~al.(2014)Henderson, Thomson, and
  Williams}]{henderson2014second}
Matthew Henderson, Blaise Thomson, and Jason~D Williams. 2014.
\newblock The second dialog state tracking challenge.
\newblock In \emph{Proceedings of the 15th Annual Meeting of the Special
  Interest Group on Discourse and Dialogue (SIGDIAL)}, pages 263--272.

\bibitem[{Henderson et~al.(2019)Henderson, Vulic, Gerz, Casanueva,
  Budzianowski, Coope, Spithourakis, Wen, Mrksic, and
  Su}]{henderson2019training}
Matthew Henderson, Ivan Vulic, Daniela Gerz, I{\~n}igo Casanueva, Pawel
  Budzianowski, Sam Coope, Georgios Spithourakis, Tsung-Hsien Wen, Nikola
  Mrksic, and Pei-Hao Su. 2019.
\newblock Training neural response selection for task-oriented dialogue
  systems.
\newblock In \emph{Proceedings of the 57th Conference of the Association for
  Computational Linguistics, ACL 2019, Florence, Italy, July 28- August 2,
  2019, Volume 1: Long Papers}, pages 5392--5404. Association for Computational
  Linguistics.

\bibitem[{Hosseini-Asl et~al.(2020)Hosseini-Asl, McCann, Wu, Yavuz, and
  Socher}]{hosseini2020simple}
Ehsan Hosseini-Asl, Bryan McCann, Chien-Sheng Wu, Semih Yavuz, and Richard
  Socher. 2020.
\newblock A simple language model for task-oriented dialogue.
\newblock \emph{arXiv preprint arXiv:2005.00796}.

\bibitem[{Kelley(1984)}]{kelley1984iterative}
John~F Kelley. 1984.
\newblock An iterative design methodology for user-friendly natural language
  office information applications.
\newblock \emph{ACM Transactions on Information Systems (TOIS)}, 2(1):26--41.

\bibitem[{Kim et~al.(2019)Kim, Kitaev, Chen, Rohrbach, Zhang, Tian, Batra, and
  Parikh}]{kim2019codraw}
Jin-Hwa Kim, Nikita Kitaev, Xinlei Chen, Marcus Rohrbach, Byoung-Tak Zhang,
  Yuandong Tian, Dhruv Batra, and Devi Parikh. 2019.
\newblock Codraw: Collaborative drawing as a testbed for grounded goal-driven
  communication.
\newblock In \emph{Proceedings of the 57th Annual Meeting of the Association
  for Computational Linguistics}, pages 6495--6513.

\bibitem[{Lan et~al.(2020)Lan, Chen, Goodman, Gimpel, Sharma, and
  Soricut}]{lan2019albert}
Zhenzhong Lan, Mingda Chen, Sebastian Goodman, Kevin Gimpel, Piyush Sharma, and
  Radu Soricut. 2020.
\newblock Albert: A lite bert for self-supervised learning of language
  representations.
\newblock In \emph{ICLR}. OpenReview.net.

\bibitem[{Lewis et~al.(2017)Lewis, Yarats, Dauphin, Parikh, and
  Batra}]{lewis2017deal}
Mike Lewis, Denis Yarats, Yann~N. Dauphin, Devi Parikh, and Dhruv Batra. 2017.
\newblock Deal or no deal? end-to-end learning of negotiation dialogues.
\newblock In \emph{Proceedings of the 2017 Conference on Empirical Methods in
  Natural Language Processing, EMNLP 2017, Copenhagen, Denmark, September 9-11,
  2017}, pages 2443--2453. Association for Computational Linguistics.

\bibitem[{Liang et~al.(2020)Liang, Tian, Chen, and Yu}]{liang2019moss}
Weixin Liang, Youzhi Tian, Chengcai Chen, and Zhou Yu. 2020.
\newblock Moss: End-to-end dialog system framework with modular supervision.
\newblock In \emph{AAAI}, pages 8327--8335. AAAI Press.

\bibitem[{Liu et~al.(2016)Liu, Lowe, Serban, Noseworthy, Charlin, and
  Pineau}]{liu2016not}
Chia-Wei Liu, Ryan Lowe, Iulian Serban, Michael Noseworthy, Laurent Charlin,
  and Joelle Pineau. 2016.
\newblock How not to evaluate your dialogue system: An empirical study of
  unsupervised evaluation metrics for dialogue response generation.
\newblock In \emph{Proceedings of the 2016 Conference on Empirical Methods in
  Natural Language Processing, EMNLP 2016, Austin, Texas, USA, November 1-4,
  2016}, pages 2122--2132. The Association for Computational Linguistics.

\bibitem[{Liu et~al.(2020)Liu, Jiang, He, Chen, Liu, Gao, and
  Han}]{liu2019radam}
Liyuan Liu, Haoming Jiang, Pengcheng He, Weizhu Chen, Xiaodong Liu, Jianfeng
  Gao, and Jiawei Han. 2020.
\newblock On the variance of the adaptive learning rate and beyond.
\newblock In \emph{Proceedings of the Eighth International Conference on
  Learning Representations (ICLR 2020)}.

\bibitem[{Liu et~al.(2019)Liu, Ott, Goyal, Du, Joshi, Chen, Levy, Lewis,
  Zettlemoyer, and Stoyanov}]{liu2019roberta}
Yinhan Liu, Myle Ott, Naman Goyal, Jingfei Du, Mandar Joshi, Danqi Chen, Omer
  Levy, Mike Lewis, Luke Zettlemoyer, and Veselin Stoyanov. 2019.
\newblock \href {http://arxiv.org/abs/1907.11692} {Roberta: A robustly
  optimized bert pretraining approach}.
\newblock \emph{CoRR}, abs/1907.11692.

\bibitem[{Ma et~al.(2019)Ma, Zeng, Zhu, Li, Yang, Yao, Zhou, and
  Shen}]{ma2019end}
Yue Ma, Zengfeng Zeng, Dawei Zhu, Xuan Li, Yiying Yang, Xiaoyuan Yao, Kaijie
  Zhou, and Jianping Shen. 2019.
\newblock \href {http://arxiv.org/abs/1912.09297} {An end-to-end dialogue state
  tracking system with machine reading comprehension and wide \& deep
  classification}.
\newblock \emph{CoRR}, abs/1912.09297.

\bibitem[{Moiseeva et~al.(2020)Moiseeva, Trautmann, and
  Sch{\"u}tze}]{moiseeva2020multipurpose}
Alena Moiseeva, Dietrich Trautmann, and Hinrich Sch{\"u}tze. 2020.
\newblock \href {http://arxiv.org/abs/2001.02284} {Multipurpose intelligent
  process automation via conversational assistant}.
\newblock \emph{CoRR}, abs/2001.02284.

\bibitem[{Peskov et~al.(2019)Peskov, Clarke, Krone, Fodor, Zhang, Youssef, and
  Diab}]{peskov2019multi}
Denis Peskov, Nancy Clarke, Jason Krone, Brigi Fodor, Yi~Zhang, Adel Youssef,
  and Mona Diab. 2019.
\newblock Multi-domain goal-oriented dialogues (multidogo): Strategies toward
  curating and annotating large scale dialogue data.
\newblock In \emph{Proceedings of the 2019 Conference on Empirical Methods in
  Natural Language Processing and the 9th International Joint Conference on
  Natural Language Processing (EMNLP-IJCNLP)}, pages 4518--4528.

\bibitem[{Rashkin et~al.(2019)Rashkin, Smith, Li, and
  Boureau}]{rashkin2018towards}
Hannah Rashkin, Eric~Michael Smith, Margaret Li, and Y-Lan Boureau. 2019.
\newblock Towards empathetic open-domain conversation models: A new benchmark
  and dataset.
\newblock In \emph{Proceedings of the 57th Conference of the Association for
  Computational Linguistics, ACL 2019, Florence, Italy, July 28- August 2,
  2019, Volume 1: Long Papers}, pages 5370--5381. Association for Computational
  Linguistics.

\bibitem[{Rastogi et~al.(2020{\natexlab{a}})Rastogi, Zang, Sunkara, Gupta, and
  Khaitan}]{rastogi2020schema}
Abhinav Rastogi, Xiaoxue Zang, Srinivas Sunkara, Raghav Gupta, and Pranav
  Khaitan. 2020{\natexlab{a}}.
\newblock Schema-guided dialogue state tracking task at dstc8.
\newblock \emph{arXiv preprint arXiv:2002.01359}.

\bibitem[{Rastogi et~al.(2020{\natexlab{b}})Rastogi, Zang, Sunkara, Gupta, and
  Khaitan}]{rastogi2019towards}
Abhinav Rastogi, Xiaoxue Zang, Srinivas Sunkara, Raghav Gupta, and Pranav
  Khaitan. 2020{\natexlab{b}}.
\newblock Towards scalable multi-domain conversational agents: The
  schema-guided dialogue dataset.
\newblock In \emph{AAAI}, pages 8689--8696. AAAI Press.

\bibitem[{Schuster and Nakajima(2012)}]{schuster2012wordpiece}
Mike Schuster and Kaisuke Nakajima. 2012.
\newblock \href {https://doi.org/10.1109/ICASSP.2012.6289079} {Japanese and
  korean voice search}.
\newblock In \emph{2012 {IEEE} International Conference on Acoustics, Speech
  and Signal Processing, {ICASSP} 2012, Kyoto, Japan, March 25-30, 2012}, pages
  5149--5152. {IEEE}.

\bibitem[{Shah et~al.(2018)Shah, Hakkani-T{\"u}r, T{\"u}r, Rastogi, Bapna,
  Nayak, and Heck}]{shah2018building}
Pararth Shah, Dilek Hakkani-T{\"u}r, G{\"o}khan T{\"u}r, Abhinav Rastogi, Ankur
  Bapna, Neha Nayak, and Larry~P. Heck. 2018.
\newblock \href {http://arxiv.org/abs/1801.04871} {Building a conversational
  agent overnight with dialogue self-play}.
\newblock \emph{CoRR}, abs/1801.04871.

\bibitem[{Suhr et~al.(2019)Suhr, Yan, Schluger, Yu, Khader, Mouallem, Zhang,
  and Artzi}]{suhr2019executing}
Alane Suhr, Claudia Yan, Jacob Schluger, Stanley Yu, Hadi Khader, Marwa
  Mouallem, Iris Zhang, and Yoav Artzi. 2019.
\newblock Executing instructions in situated collaborative interactions.
\newblock In \emph{Proceedings of the 2019 Conference on Empirical Methods in
  Natural Language Processing and the 9th International Joint Conference on
  Natural Language Processing, EMNLP-IJCNLP 2019, Hong Kong, China, November
  3-7, 2019}, pages 2119--2130. Association for Computational Linguistics.

\bibitem[{Weston et~al.(2016)Weston, Bordes, Chopra, and
  Mikolov}]{weston2015babi}
Jason Weston, Antoine Bordes, Sumit Chopra, and Tomas Mikolov. 2016.
\newblock \href {http://arxiv.org/abs/1502.05698} {Towards ai-complete question
  answering: A set of prerequisite toy tasks}.
\newblock In \emph{4th International Conference on Learning Representations,
  ICLR 2016, San Juan, Puerto Rico, May 2-4, 2016, Conference Track
  Proceedings}.

\bibitem[{Williams et~al.(2017)Williams, Asadi, and Zweig}]{williams2017hybrid}
Jason~D. Williams, Kavosh Asadi, and Geoffrey Zweig. 2017.
\newblock Hybrid code networks: practical and efficient end-to-end dialog
  control with supervised and reinforcement learning.
\newblock In \emph{Proceedings of the 55th Annual Meeting of the Association
  for Computational Linguistics, ACL 2017, Vancouver, Canada, July 30 - August
  4, Volume 1: Long Papers}, pages 665--677. Association for Computational
  Linguistics.

\bibitem[{Wolf et~al.(2019)Wolf, Sanh, Chaumond, and
  Delangue}]{wolf2019transfertransfo}
Thomas Wolf, Victor Sanh, Julien Chaumond, and Clement Delangue. 2019.
\newblock \href {http://arxiv.org/abs/1901.08149} {Transfertransfo: A transfer
  learning approach for neural network based conversational agents}.
\newblock \emph{CoRR}, abs/1901.08149.

\bibitem[{Wu et~al.(2019)Wu, Madotto, Hosseini-Asl, Xiong, Socher, and
  Fung}]{wu2019transferable}
Chien-Sheng Wu, Andrea Madotto, Ehsan Hosseini-Asl, Caiming Xiong, Richard
  Socher, and Pascale Fung. 2019.
\newblock Transferable multi-domain state generator for task-oriented dialogue
  systems.
\newblock In \emph{Proceedings of the 57th Conference of the Association for
  Computational Linguistics, ACL 2019, Florence, Italy, July 28- August 2,
  2019, Volume 1: Long Papers}, pages 808--819. Association for Computational
  Linguistics.

\bibitem[{Wu and Xiong(2020)}]{wu2020probing}
Chien-Sheng Wu and Caiming Xiong. 2020.
\newblock \href {https://www.aclweb.org/anthology/2020.emnlp-main.409} {Probing
  task-oriented dialogue representation from language models}.
\newblock In \emph{Proceedings of the 2020 Conference on Empirical Methods in
  Natural Language Processing, Virtual Online, November 16 - November 20,
  2020}, pages 5036--5051, Online. Association for Computational Linguistics.

\bibitem[{Zhang et~al.(2018)Zhang, Dinan, Urbanek, Szlam, Kiela, and
  Weston}]{zhang2018personalizing}
Saizheng Zhang, Emily Dinan, Jack Urbanek, Arthur Szlam, Douwe Kiela, and Jason
  Weston. 2018.
\newblock Personalizing dialogue agents: I have a dog, do you have pets too?
\newblock In \emph{Proceedings of the 56th Annual Meeting of the Association
  for Computational Linguistics, ACL 2018, Melbourne, Australia, July 15-20,
  2018, Volume 1: Long Papers}, pages 2204--2213. Association for Computational
  Linguistics.

\bibitem[{Zhang et~al.(2020)Zhang, Sun, Galley, Chen, Brockett, Gao, Gao, Liu,
  and Dolan}]{zhang2019dialogpt}
Yizhe Zhang, Siqi Sun, Michel Galley, Yen-Chun Chen, Chris Brockett, Xiang Gao,
  Jianfeng Gao, Jingjing Liu, and Bill Dolan. 2020.
\newblock Dialogpt: Large-scale generative pre-training for conversational
  response generation.
\newblock In \emph{Proceedings of the 58th Annual Meeting of the Association
  for Computational Linguistics: System Demonstrations, ACL 2020, Online, July
  5-10, 2020}, pages 270--278. Association for Computational Linguistics.

\bibitem[{Zhou et~al.(2018)Zhou, Young, Huang, Zhao, Xu, and
  0001}]{zhou2018commonsense}
Hao Zhou, Tom Young, Minlie Huang, Haizhou Zhao, Jingfang Xu, and Xiaoyan~Zhu
  0001. 2018.
\newblock \href {http://www.ijcai.org/proceedings/2018/} {Commonsense knowledge
  aware conversation generation with graph attention}.
\newblock In \emph{Proceedings of the Twenty-Seventh International Joint
  Conference on Artificial Intelligence, IJCAI 2018, July 13-19, 2018,
  Stockholm, Sweden}, pages 4623--4629. ijcai.org.

\end{thebibliography}
\bibliographystyle{acl_natbib}

\clearpage
\appendix
\section{Agent Training Details}
\label{sec:training}
Optimizing agents performance can be split into preparation before the HIT (Human Intelligence Task), improving HIT itself, and ongoing training afterwards. Starting with the pre-HIT phase, the major steps largely center around multiple rounds of qualifications to filter for the highest quality workers available.  During the post-HIT phase, effort shifts to ensuring that each worker becomes increasingly comfortable with the task.

\paragraph{Pre-HIT Phase} Qualifications take the form of online quizzes which serve the purpose of training motivated workers in addition to simply removing unqualified ones.
When designing the qualification, the number and style of questions were iterated on to limit the feeling of a tight time constraint, while still remaining quite difficult. In fact, some agents who had previously had actual customer service jobs mentioned they felt like they were right back at the office. This difficulty resulted in a high rejection rate, which was costly because we paid Turkers \$2 regardless of passing the exam (with a larger \$8 bonus for passing). Although, the cost was well worth the trade-off since having high quality agents would pay dividends down the road.

To move efficiently, we leaned heavily on multiple choice questions and MTurk APIs to help automate grading and assignment of qualifications. Finally, we learned that including screenshots of the Agent Dashboard in the quizzes was a great way to familiarize the agents with the platform before performing the actual task.

\paragraph{During-HIT Phase} The HIT itself was priced at \$1.50 for completing the conversation with an extra \$1.00 bonus for identifying the correct customer intent at the end-of-chat survey.  Since agents are naturally focused on getting done as quickly as possible, they would often only take the customer's requests into account, bypassing a key characteristic of what makes ABCD unique. However, by encouraging agents to focus on the customer intent, they were forced to peruse the Agent Guidelines for the associated subflow. Thus, we found this incentive critical for aligning agent behaviors with optimal outcomes.

\paragraph{Post-HIT Phase} For ongoing training, we began producing small lists of bulletpoints to the agents on areas they could improve on.  Furthermore, we would highlight examples of good and bad decision-making and appropriate behavior when representing the fictitious ``AcmeBrands'' retail company.  Finally, we also recorded videos which gave agents a view of how an ``ideal'' agent would behave at every step of the chat.  We found that by engaging with the Turkers through the group chat and respecting their feedback, they were very willing to work on improving despite not having extra monetary incentive to do so.  

\begin{figure}
  \centering
  \includegraphics[width=1.04\linewidth]{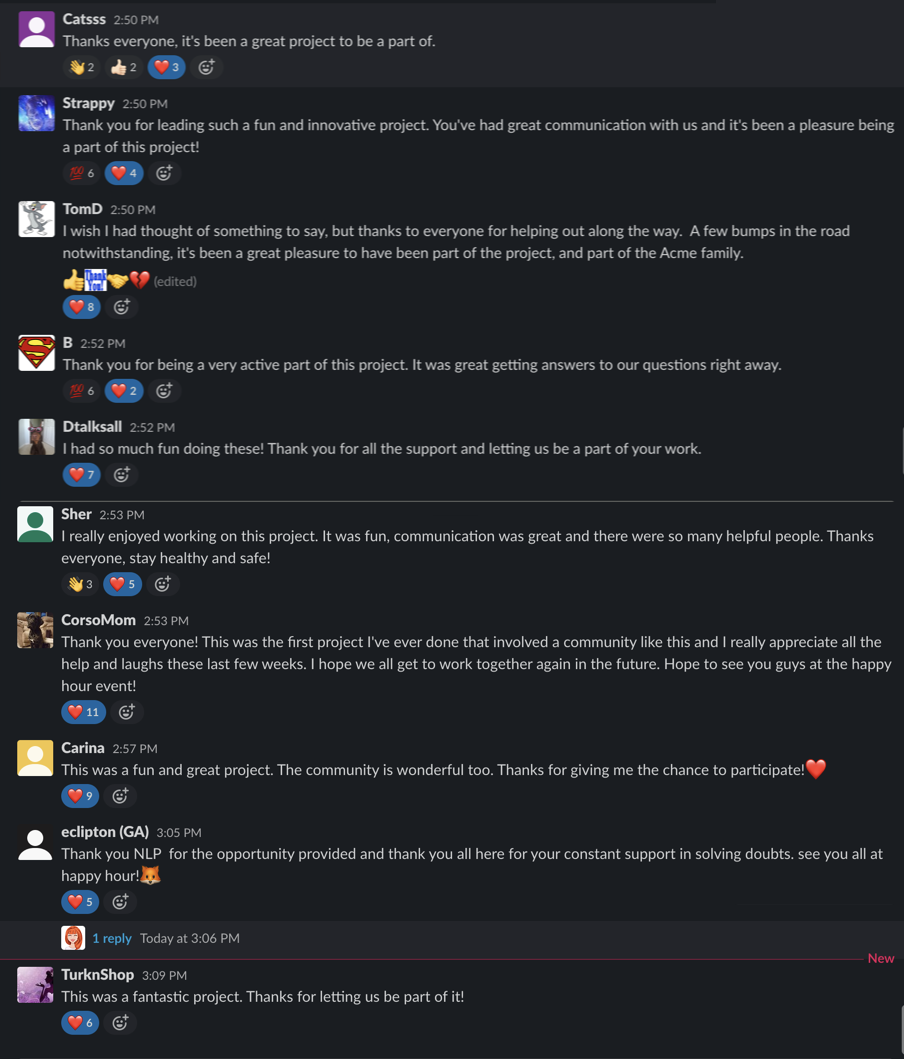}
  \caption{Crowdworker feedback in chat platform after the completion of the final batch of data collection.}
  \label{fig:feedback}
\end{figure}

In total, the agents were quite wonderful to work with and their end-of-task feedback strongly suggests they enjoyed the process as well. (See Figure~\ref{fig:feedback})  We credit this to the training details mentioned in this section and the development of the \livechat~procedure.

\vfill\null
\pagebreak

\section{Optimizing Available Workers}
\label{sec:pairing}
In a regular Mechanical Turk (MTurk) setup, HITs are made available to a large audience who can pick up as many or as few as they want.  \livechat~dictates a dialogue between two speakers, so we need two types of workers: agents and customers.  Let us consider the number of agents available as $A$ and the number of customers available as $C$.  Given budget constraints, we can only pay some maximum number of workers $M$.  Simultaneously, given time constraints, we need a minimum number of conversations collected per week, which is a function of the number of available workers $N = f(A,C)$. This leads to three issues that must be considered in conjunction:

\paragraph{N $<$ A+C $<$ M} Operating the Agent Dashboard requires a highly skilled worker, so efficient data collection is limited by the number of available agents.  Although the customer side of ABCD is a simpler task, there is still a minimum bar to be met to prevent (a) customers who spam with random text (b) customers who fake scenarios or (c) customers who hoard HITs and never show up to the chat.  Thus, there needs to be a sufficient amount of both agent $A$ and customers $C$ qualified and available in order to surpass the minimum threshold set by $N$.  However, simply paying more per HIT bumps up against the limits set by $M$.

\paragraph{C $>>$ A} Since training agents is more resource intensive than training customers, it makes sense to simply have more of the latter.  Yet by doing so leads to an issue where customers wait around for agents when they arrive in the waitroom.  In a typical scenario, a customer might leave the tab open to work on other tasks, but when they are eventually paired, the customer is often busy doing something else, leaving the chat to flounder.  In the worst case, the customer starts to verbally abuse the agent about the long wait time when they are finally paired.

\paragraph{A $>>$ C} Finding as many agents as possible is not the solution either because now the agents will end up waiting around for customers.  If the waiting periods are too long, agents will abandon the task and disparage your reputation on various forms of social media.  Since the task is difficult, the pool of workers who may eventually qualify as agents is finite, so too many poor interactions can halt the data collection process completely. 

To resolve this situation, we begin with the maximum number of workers $M$ as the starting constraint given a fixed budget. If we qualify too many workers, then we will not have enough budget left for the actual conversations, so instead we qualify workers in mini-batches.  Since the pool of potential workers who may meet the strict requirements for agents $A$ is more limited than customer candidates $C$, we start on the agent side.  Given some amount of qualified agents, only a percentage of them will show up at the desired time slot to perform the task.  Thus, we increment the batch size until the number of available workers passes the minimum $A > N/2$.

To limit the number of customers who show up, we filter for users by location, number of completed HITS, and sufficient rating.  We also establish an exam that is purposely very easy (to minimize costs), but just hard enough to deter bots and spammers. To raise the likelihood that the customer will show up, we include a question in the quiz which simply asks when the customer is available to perform the HIT.  We really emphasize this question and make it required, so workers are aware of its significance.  This allows us to tune the customer count such that $C \approx A$.

Note that due to the higher pay rate, agents are more likely to show up than customers.  Therefore, there needs to be a higher ratio of customers to account for this imbalance.  For some intuition on where to start, we found that a good rule of thumb was to consider the appearance ratio as inversely proportional to the ratio of pay. One final insight is to make the HITs heavily dependent on bonus pay, with base pay very low.  This will keep spammers away since they will end up with a pittance when attempting to game the system.

To improve the waitroom experience, we added a feature where a user's place in the queue would be updated live, along with a timer indicating the expected wait.  For Turkers willing to wait around, helpful and encouraging messages would also be displayed to keep them occupied.  Alternatively, for Turkers who were multi-tasking, visual and audio notifications were added to signify the start of a chat, allowing them to attend to other tasks in the meantime. We believe our modifications have only scratched the surface and that improving the user experience for data collection offers an interesting line of HCI research to explore.

\vfill\null

\section{Cascading Evaluation}
\label{sec:cascade}
To motivate cascading dialogue success (CDS) over typical other accuracy metrics, consider the scenario where a model gets 80\% of turns correct, while still achieving 0\% accuracy on the conversation level because it always messes up somewhere right at the end of the dialogue.  A turn-based metric would over-estimate performance since such a metric fails to capture the model’s consistent shortcomings in closing conversations.  On the other hand, conversation-based metrics under-estimate the model's performance because such measures fail to account for the fact that the system is mostly successful.  Moreover, each evaluation would be limited to occurring only once per conversation, which makes inefficient use of scarce data as a resource.

Instead, cascading dialogue success creates an evaluation example for the remainder of each conversation starting from each turn.  For example, suppose a chat contained 4 turns: [A, B, C, and D], training instances can be created with this data that include: [A, B, C, D], [B, C, D], [C, D] and [D] by itself.  Now imagine the model consistently predicted turn $C$ incorrectly, and everything else correct.  Then its scores would be $2/4$, $1/3$, $0$ and $1$, respectively. Averaging across all turns would yield a final cascading success rate of 45.8\%.  A turn-based metric would yield 75\% while a conversation-based metric would yield 0\%.  Thus, CDS allows a model to earn partial credit on what it has learned without severe penalties in either direction.

\section{Model Training Hyperparameters}
\label{sec:params}
When training the best model for Action State Tracking, we ended up with a learning rate of 3e-5, hidden dimension of 1024, weight decay of 0.05 and a batch size of 10 examples.  Training lasted for 14 epochs, where we early stopped if overall accuracy failed to improve for three epochs in a row.  The RAdam optimizer had a linear warm-up for three epochs, with hyper-parameters kept at their defaults of 0.9 and 0.999.  We also add the delexicalized slots into the vocabulary of the tokenizer.

For Cascading Dialogue Success, our best model had a 1e-5 learning rate, 1024 hidden dimension and no weight decay.  The batch size was shrunk to 3 examples, but this was due purely to memory rather than performance reasons.  Training was set to 21 epochs, and again we early stopped if overall accuracy failed to improve for three epochs in a row.  Finally, the optimizer again had a linear warm-up for three epochs with hyper-parameters kept at their defaults.

\section{Intent Info and Guidelines}
\label{sec:info}
We augment the model with access to intent information in two ways.  First, the subflow is translated into an index which is concatenated to all input contexts so the model can leverage this information.  Second, the intent classifier is directly fed the solution, which is what allows it to trivially reach perfect accuracy.

We leverage the Agent Guidelines by using it to mask invalid action predictions.  More specifically, given a predicted subflow, the guidelines outline all possible actions and values within that subflow.   With this information, a mask is created before training and applied during evaluation to only allow valid actions.  

\section{Conversation Examples}
Since ABCD was collected using \livechat~rather than templates, we observe various linguistic diversity in the chats.  These phenomena limit the ability of models to memorize artificial patterns when making predictions.

\begin{table}[H]
\centering
\resizebox{\linewidth}{!}{%
\begin{tabular}{l@{\hspace{0.5em}}}
\hline
\textbf{Co-reference} \\
\textsc{CUS}: I'd like to return something \\
\textsc{AGT}: OK \\
\textsc{AGT}: Can I get your full name \\
\textsc{AGT}: Also user name, email address, order id \\
\textsc{AGT}: Membershp level and reason for return \\
\textsc{CUS}: Alessandro Phoenix, aphoenix872@email.com, order ID is 4024067912 \\
\textsc{CUS}: I'm at the Gold level. I'm returning it because it's the wrong size \\
\hline
\textbf{Chit-Chat} \\
\textsc{AGT}: Do you need any more help? \\
\textsc{CUS}: a break, I need a coffee break~ \\
\textsc{CUS}: but no, nothing from you \\
\textsc{CUS}: thanks for the save \\
\textsc{AGT}: Haha have a good break! And have an even better day. \\
\hline
\textbf{Emotion} \\
\textsc{AGT}: Ok, there was a mistake made.  Do you have the Shipping Status? \\
\textsc{CUS}: It's in transit \\
\textsc{AGT}: Ok, that means it's already out for shipment\\
\textsc{CUS}: so two are being sent?\\
\textsc{AGT}: Yes.  Unfortunately that means when you get the item you will \\ \quad need to call back and make a return \\
\textsc{CUS}: oh you gotta be kidding me! \\
\hline
\end{tabular}}
\caption{Examples of linguistic phenomenon.} \label{tab:lang-phenom}
\end{table}

\section{Agent Guidelines}
\label{sec:agent}
\textit{(screenshot on following page)}

\begin{figure*}
  \centering
  \includegraphics[width=0.7\linewidth]{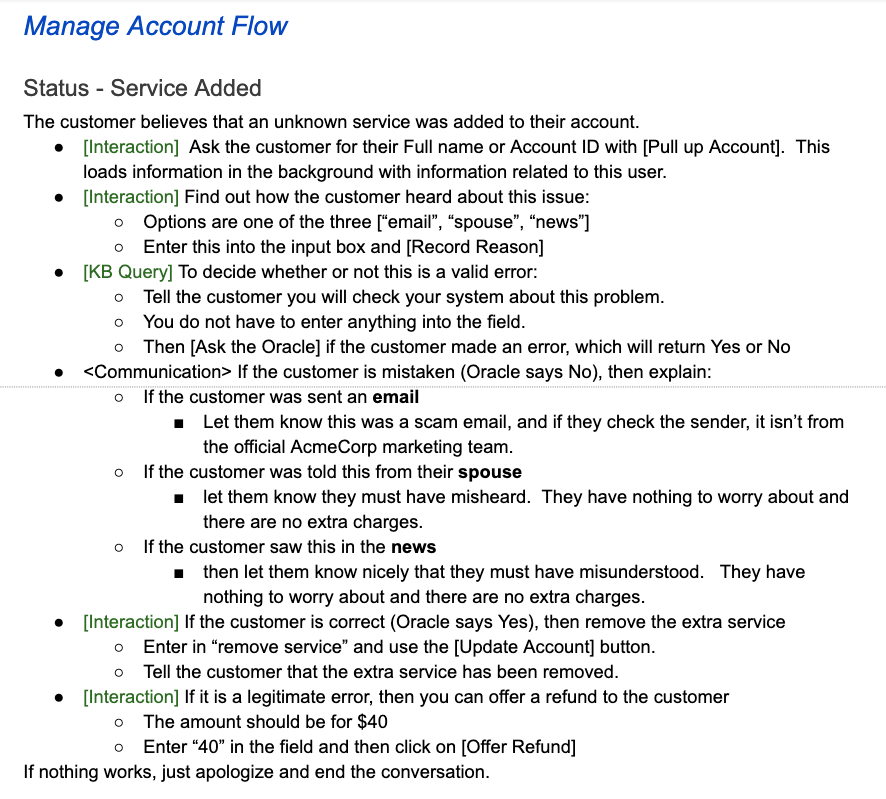}
  \caption{A example subflow \textit{Status - Service Added} under the \textit{Mange Account Flow} in the Agent Guidelines}
  \label{fig:agent_guidelines}
\end{figure*}

\section{Customer Panel}
\label{sec:customer}
\textit{(screenshot on following page)}

\begin{figure*}
  \centering
  \includegraphics[width=0.7\linewidth]{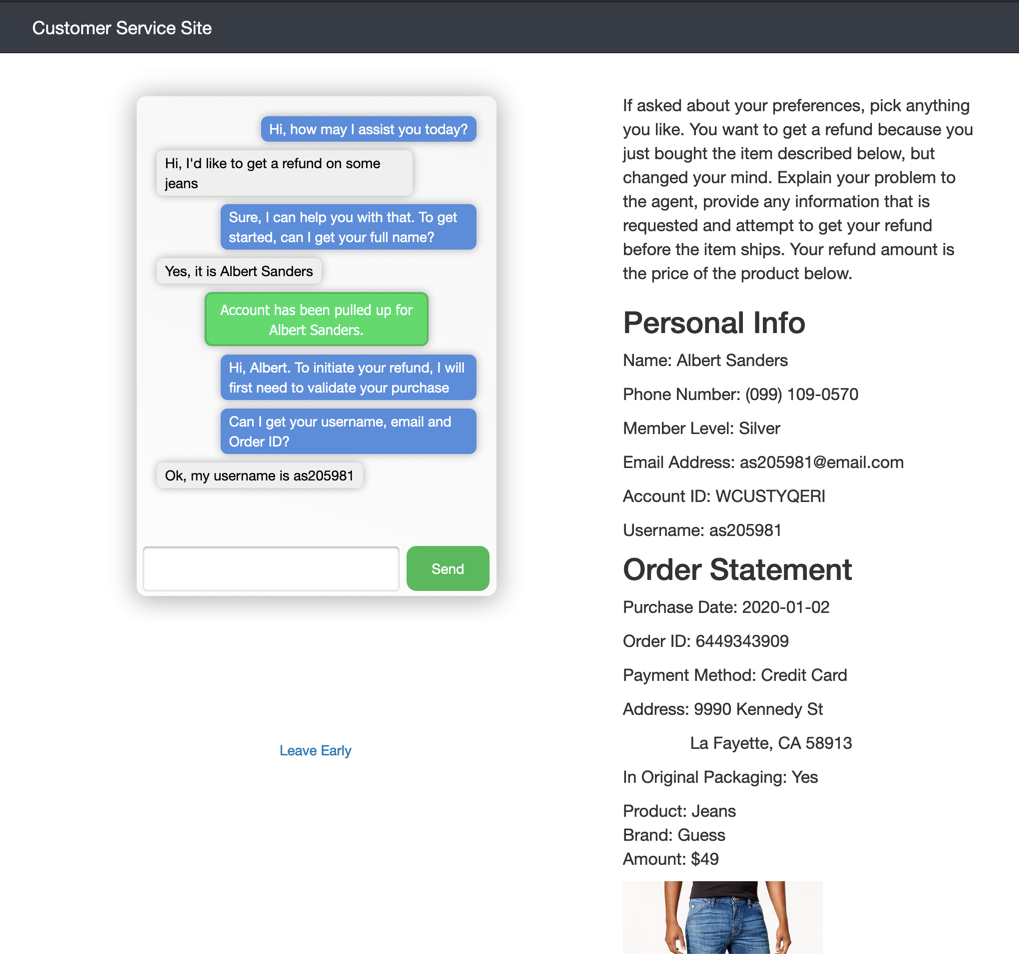}
  \caption{The customer chat interface (left) shows an on-going conversation with customer messages in grey, agent messages in blue, and actions in green. The customer prompt (right top) grounds the customer to a specific issue and backstory. The info sections (right middle and bottom) contains values that the customer has to provide in the conversation as well as other meta-data such as product information.}
  \label{fig:customer_interface}
\end{figure*}
\end{document}